\documentclass[11pt]{article}
\usepackage{acl2016}
\usepackage{times}
\usepackage{latexsym}
\usepackage{multirow}
\usepackage{arydshln}
\usepackage{amssymb,amsfonts}
\usepackage{amsmath}
\usepackage{rotating}
\usepackage{color}
\usepackage{subfigure}
\usepackage{url}
\aclfinalcopy 
\def\aclpaperid{***} 

\def\figref#1{Figure~\ref{fig:#1}}
\def\figlabel#1{\label{fig:#1}\label{p:#1}}

\def\tabref#1{Table~\ref{tab:#1}}
\def\tablabel#1{\label{tab:#1}\label{p:#1}}

\def\secref#1{Section~\ref{sec:#1}}
\def\seclabel#1{\label{sec:#1}\label{p:#1}}
\def\eqref#1{Eq.~\ref{eqn:#1}}

\long\def\eat#1{\ignorespaces}

\title{Why and How to Pay Different Attention to Phrase Alignments\\ of Different Intensities}

\author{Wenpeng Yin \rm{and} \textbf{Hinrich Sch\"{u}tze}\\
	    Center for Information and Language Processing\\University of Munich, Germany\\
	    {\tt wenpeng@cis.uni-muenchen.de}}\author{Wenpeng Yin \rm{and} \textbf{Hinrich Sch\"{u}tze}\\
	    Center for Information and Language Processing\\LMU Munich, Germany\\
	    {\tt wenpeng@cis.lmu.de}}

\date{}

\newcounter{notecounter}
\newcommand{\enotesoff}{\long\gdef\enote##1##2{}}
\newcommand{\enoteson}{\long\gdef\enote##1##2{{
\stepcounter{notecounter}
\large\bf
\hspace{1cm}\arabic{notecounter} $<<<$ ##1: ##2
$>>>$\hspace{1cm}}}}
\enoteson
\enotesoff

\begin{document}

\maketitle

\begin{abstract}
This work studies comparatively  two typical sentence pair classification tasks: textual entailment (TE) and answer selection (AS), observing that phrase alignments of different intensities contribute differently in these tasks. We address the problems of identifying
phrase alignments of \emph{flexible granularity} and pooling \emph{different
intensities\footnote{By intensity we roughly mean the degree of match.}} for these tasks. 
Prior
work  (i) has limitations in phrase generation and representation, or (ii)
conducts alignment at word and phrase levels by
handcrafted features or (iii) utilizes a single framework of
alignment
without considering the characteristics of
specific tasks, which limits the framework's effectiveness
across tasks. We propose an architecture based on Gated Recurrent Unit that supports
(i) representation learning of phrases of
\emph{arbitrary granularity} and (ii) task-specific phrase
alignment between two sentences by \emph{attention
  pooling}. 
Experimental results on TE and AS match our observation and are state-of-the-art.
\end{abstract}

\enote{hs}{The abstract still has the problem that you
  define intensity, but you do not say why intensity is
  important and how you use it.

if intensity is an important contribution (and the fact you
mention it in the title and define it in the abstract
suggests it is an important contribution), then you should say
in the abstract why it is an important contribution\

i assume that attention pooling is based on intensity? at
this point the reader will ask: what is attention pooling?
what's so great about intensity?

}


\section{Introduction}
\seclabel{intro}
How to model a pair of sentences is a critical issue in many
NLP tasks, including textual
entailment 
\cite{marelli2014semeval,bowman2015large,yin2015abcnn} and
answer selection
\cite{yu2014deep,yang2015wikiqa,santos2016attentive}.
A key challenge common to  these tasks is the
lack of explicit alignment annotation between the sentences
of the pair.
Thus, inferring and assessing the semantic relations
between words and phrases in the two sentences is a core issue.

\begin{figure}[t]
\setlength{\belowcaptionskip}{-10pt}
\setlength{\abovecaptionskip}{0pt}
\centering
\includegraphics[width=7.5cm]{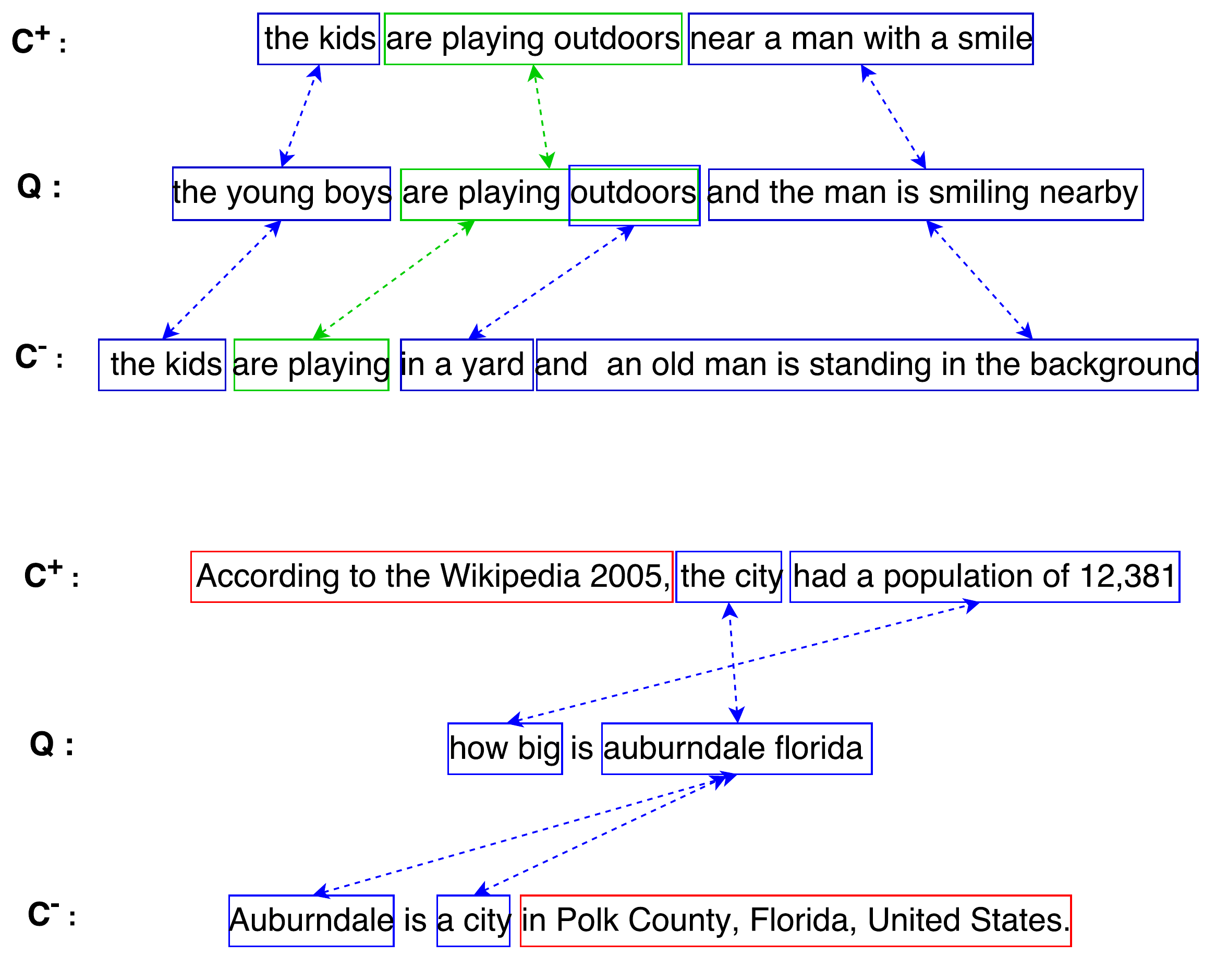}
\caption{Alignment examples in TE (top) and AS
  (bottom). Green color: identical (subset) alignment; blue color: relatedness alignment; red color: unrelated alignment. Q: the first sentence in TE or the question in AS; $C^+$, $C^-$: the correct or incorrect counterpart in the sentence pair ($Q$, $C$).} \label{fig:example}
\end{figure}

Figure \ref{fig:example} shows examples of alignments.  In
the TE example, we try to figure out $Q$ entails $C^+$
(positive) or $C^-$ (negative).  As human beings, we
discover the relationship of two sentences by studying the
alignments between linguistic units. We see that some
phrases are kept: ``are playing outdoors'' (between $Q$ and
$C^+$), ``are playing '' (between $Q$ and $C^-$). Some phrases
are changed into related semantics on purpose: ``the young
boys'' ($Q$) $\rightarrow$ ``the kids'' ($C^+$ \& $C^-$),
``the man is smiling nearby'' ($Q$) $\rightarrow$ ``near a
man with a smile'' ($C^+$) or $\rightarrow$ ``an old man is
standing in the background'' ($C^-$) . We can see that the
kept parts have strong alignments (green color), and changed
parts have weaker alignments (blue color). To
successfully identify the relationships of ($Q$, $C^+$) or
($Q$, $C^-$), studying the changed parts is
crucial. \emph{Hence, we argue that TE should pay more
  attention to weaker alignments. }

In AS, we try to figure out: does sentence $C^+$ or sentence
$C^-$ answer question $Q$?  Roughly, the content in
candidates $C^+$ and $C^-$ can be classified into aligned
part (e.g., repeated or relevant parts) and indifferent
part. This differs from TE, in which it is hard to claim
that some parts are indifferent, as TE requires to make
clear that each part can entail or be entailed. Hence, TE is
considerably sensitive to those ``unseen'' parts. In
contrast, \emph{AS is more tolerant of indifferent parts and
  less related parts.}  From the AS example in Figure
\ref{fig:example}, we see that ``Auburndale Florida'' ($Q$)
can find related part ``the city'' ($C^+$), and
``Auburndale'', ``a city'' ($C^-$) ; ``how big'' ($Q$) also
matches ``had a population of 12,381'' ($C^+$) very
well. And some unaligned parts exist, denoted by red
color. \emph{Hence, we argue that AS needs to pay more
  attention to stronger alignments.}

The above analysis suggests that: (i) alignments
connecting two sentences can happen between phrases of
arbitrary granularity; (ii) phrase alignments can have
different intensities; (iii) tasks of different properties
require paying different attention to alignments of
different intensities.

Alignments at word level \cite{yih2013question} or phrase
level \cite{yao2013semi} both have been studied before.  For
example, \newcite{yih2013question} make use of WordNet
\cite{Miller95} and Probase \cite{wu2012probase} for
identifying hypernymy, hyponymy. \newcite{yao2013semi} use
POS tags, WordNet and paraphrase database for alignment
identification. Their approaches  rely on
manual feature design and linguistic resources. We develop a deep
neural network (DNN) to learn representations of phrases of
arbitrary lengths. As a result, alignments can be searched
in a more exhaustive way.

DNNs have been intensively investigated in
sentence pair classifications
\cite{blacoe2012comparison,socher2011dynamic,yin2015ACL},
and attention mechanisms are also applied to individual
tasks
\cite{santos2016attentive,entail2016,wang2015learning};
however, most attention-based DNNs have implicit assumption
that stronger alignments deserve more attention
\cite{yin2015abcnn,santos2016attentive}. Our examples in
Figure \ref{fig:example}, instead, show that this assumption
does not hold invariably. Weaker alignments in certain tasks
such as TE can be the indicator of the final
decision. This motivates us in this work to introduce  DNNs
with a flexible
attention mechanism that is adaptable for specific tasks. For TE, it can make
our system pay more attention to weaker alignments; for
AS, it enables our system to focus on stronger
alignments. In experiments, we will show that this attention
scheme is very effective for different tasks.

We make the following contributions. (i) 
We use GRU
(Gated Recurrent Unit \cite{cho2014properties}) to learn representations for phrases of arbitrary
granularity. Based on  phrase representations, we can detect phrase alignments
of different intensities. (ii) We propose attention pooling 
to achieve flexible choice among alignments, depending on
the characteristics of the task. (iii) We achieve
state-of-the-art
performance on TE and AS.



\section{Related Work}
\textbf{Non-DNN for sentence pair modeling.}
\newcite{heilman2010tree} describe tree edit models that
generalize tree edit distance by allowing operations that
better account for complex reordering phenomena and by
learning from data how different edits should affect the
model's decisions about sentence
relations. \newcite{wang2010probabilistic} cope with the
alignment between a sentence pair by using a probabilistic
model that models tree-edit operations on dependency parse
trees. Their model treats alignments as structured latent
variables, and offers a principled framework for
incorporating complex linguistic features.
\newcite{yih2013question} try to improve the shallow
semantic component, lexical semantics, by formulating
sentence pair as a semantic matching problem with a latent
word-alignment structure as in
\cite{chang2010discriminative}. More fine-grained word
overlap and alignment between two sentences are explored in
\cite{lai2014illinois}, in which negation, hypernym/hyponym,
synonym and antonym relations are
used. \newcite{yao2013semi} extend word-to-word
alignment to phrase-to-phrase alignment by a semi-Markov
CRF.

\textbf{DNN for sentence pair classification.} There
recently has been great interest in using DNNs for
classifying sentence pairs as they can reduce the burden of feature engineering.

For TE, \newcite{bowman2015recursive} employ recursive
DNN to encode entailment on SICK
\cite{marelli2014sick}.  \newcite{entail2016}
present an attention-based LSTM (long short-term memory,
\newcite{hochreiter1997long}) for the SNLI corpus \cite{bowman2015large}. 

For AS, \newcite{yu2014deep} present a bigram CNN (convolutional neural network) to model
question and answer candidates. 
\newcite{yang2015wikiqa} extend this method 
and get state-of-the-art performance on the
WikiQA dataset. 
\newcite{feng2015applying} test various setups of a
bi-CNN architecture on an insurance domain QA dataset.
\newcite{tan2015lstm} explore
bidirectional LSTM on the same
dataset. Other sentence pair classification tasks 
including paraphrase identification \cite{socher2011dynamic,yinnaacl}
are also investigated.

Some prior work aims to solve a general sentence matching
problem, e.g., \newcite{hu2014convolutional},\newcite{yin2015ACL},\newcite{wan2015deep}.


\textbf{Attention-based DNN for alignment.} DNNs have been
successfully developed to detect alignments, e.g., in
machine translation
\cite{bahdanau2015neural,luong2015effective} and text
reconstruction \cite{li2015hierarchical,rush2015neural}.
In addition, attention-based alignment is also applied
in natural language inference (e.g.,
\newcite{entail2016},\newcite{wang2015learning}, \newcite{cheng2016long}). However,
most of this work aligns word-by-word. As Figure
\ref{fig:example} shows, many sentence relations can be better
identified through phrase level alignments. This  is one
motivation of our work.

\section{Model}
This section first gives a brief introduction of GRU and how it performs phrase representation learning, then describes the different attention poolings for phrase alignments w.r.t TE and AS tasks.


\subsection{GRU Introduction}
GRU  is a
simplified version of LSTM. Both are found effective in sequence modeling, as they are order-sensitive and can capture long-range context.  The
tradeoffs between GRU and its competitor LSTM have not been
fully explored yet. According to empirical evaluations in
\cite{chung2014empirical,jozefowicz2015empirical}, there is
not a clear winner. In many tasks both architectures yield
comparable performance and tuning hyperparameters like layer
size is probably more important than picking the ideal
architecture. GRU have fewer parameters and thus may train a
bit faster or need less data to generalize. Hence, we use
GRU to model text:
\begin{eqnarray}\label{equ:gru}
\setlength{\abovedisplayskip}{-8pt}
\setlength{\belowdisplayskip}{3pt}
z&=&\sigma(x_tU^z+s_{t-1}W^z)\\
r&=&\sigma(x_tU^r+s_{t-1}W^r)\\
h_t&=&\mathrm{tanh}(x_tU^h+(s_{t-1}\circ r)W^h)\\
s_t&=&(1-z)\circ h_t+z\circ s_{t-1}
\end{eqnarray}
$x$ is the input sequence with word $x_t$ at  position $t$,
$s_t$ is the hidden state at  $t$, $U$ and $W$ are parameters.

\enote{hs}{check: below: deleted this}



\subsection{Representation Learning for Phrases}\label{sec:rep}
\begin{figure}[t] 
\setlength{\belowcaptionskip}{-10pt}
\setlength{\abovecaptionskip}{0pt}
\centering 
\begin{tabular}{cc}
\includegraphics[width=3.5cm]{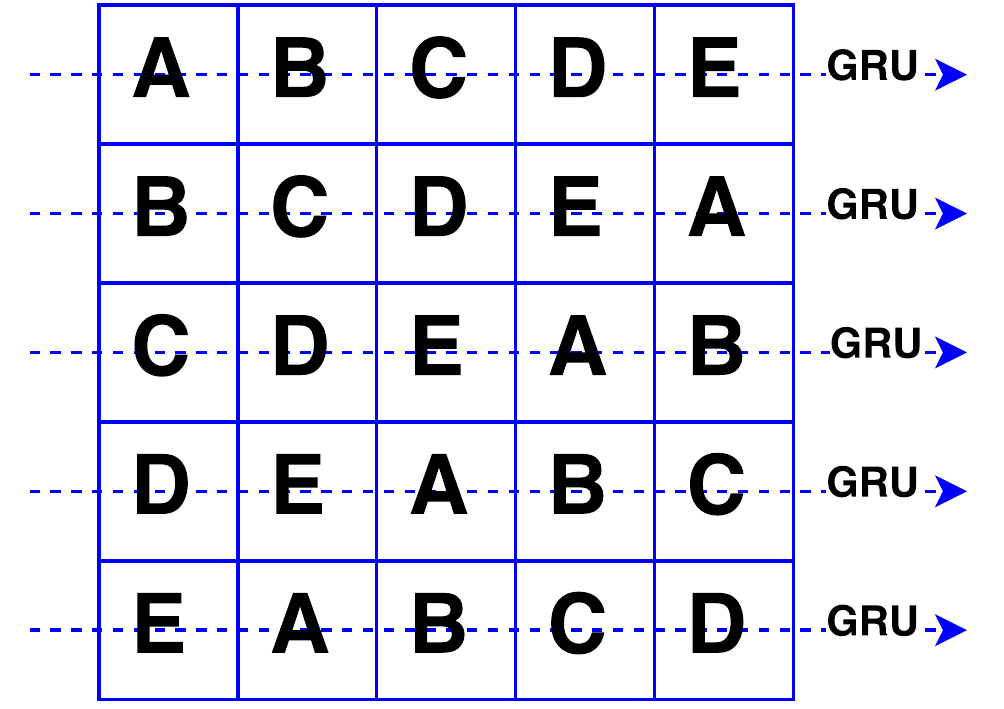} 
&\includegraphics[width=3.7cm]{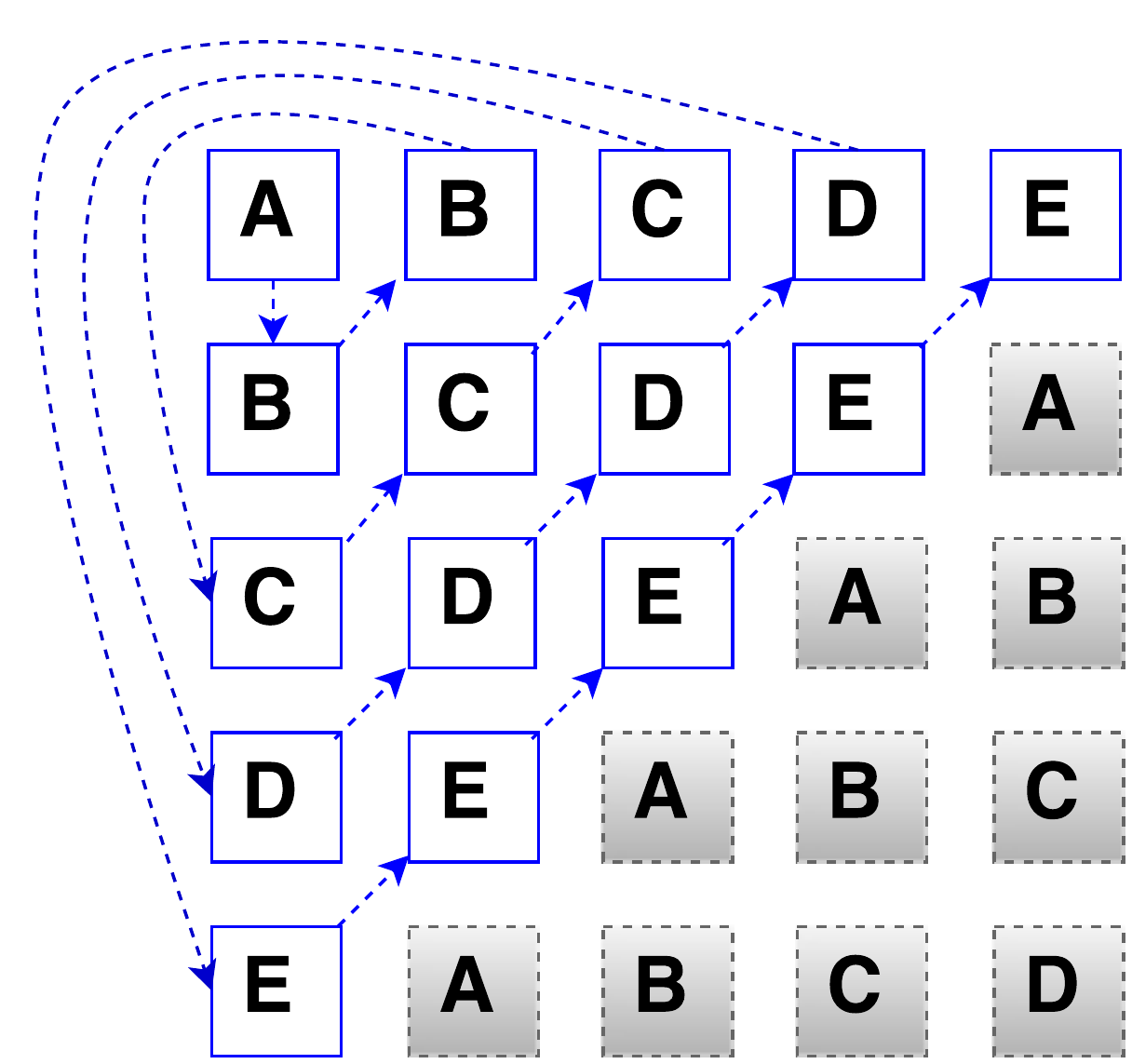} 
\end{tabular}
\caption{Phrase representation learning by GRU (left), sentence reformatting by connecting GRU hidden states, i.e, phrase representations (right)} 
\label{fig:rep} 
\end{figure}

For a general sentence $s$ with five consecutive words:
ABCDE, with each word represented by a word embedding of
dimensionality $d$, we first create four fake sentences,
$s^1$: ``BCDEA'', $s^2$: ``CDEAB'', $s^3$: ``DEABC'' and
$s^4$: ``EABCD'', then put them in a tensor of size $5\times 5\times d$
(\figref{rep}, left). Note that each character in \figref{rep} (left) denotes an embedding, we neglect its dimensionality to make the figure simpler (looks like a matrix).

We run a same GRU on each row of this matrix in parallel. As GRU is able to
encode the whole sequence up to current position, this step
generates representations for any consecutive phrases in
original sentence $s$. For example, the GRU hidden state at
position ``E'' at coordinates (1,5) (i.e., 1st row, 5th
column) denotes the representation of the phrase ``ABCDE''
which in fact is $s$ itself, the hidden state at ``E'' (2,4)
denotes the representation of phrase ``BCDE'', $\ldots$ ,
the hidden state of ``E'' (5,1) denotes phrase
representation of ``E'' itself. \emph{Hence, for each token,
  we can learn the representations for all phrases ending
  with this token}. Finally, all phrases of any lengths in
$s$ can get a representation vector. GRUs in those rows are
set to share weights so that all phrase representations are
comparable in the same space.

Now, we reformat sentence ``ABCDE'' into $s^*$ = ``(A) (B) (AB) (C)
(BC) (ABC) (D) (CD) (BCD) (ABCD) (E) (DE) (CDE) (BCDE)
(ABCDE)'' by connecting corresponding GRU hidden states, as shown by arrows in \figref{rep} (right) (note that each character in \figref{rep} (right) \emph{denotes the hidden states} generated by GRUs in \figref{rep} (left)).
Each
sequence in parentheses is a phrase (we use parentheses just
for making the phrase boundaries clear). Randomly taking a
phrase ``CDE'' as an example, its representation  comes
from the hidden state at ``E'' (3,3) in
\figref{rep} (left). Shaded parts are discarded. The main advantage of reformatting
sentence ``ABCDE'' into the \emph{new sentence} $s^*$ is to
create phrase-level semantic units, but at the same time we maintain
the order information. As a result, the initial order-3 tensor input will generate a new matrix representing the sentence $s^*$ after GRU operation by putting those phrase representations in sequence, e.g, in columns sequentially.

Hence, the sentence ``how big is Auburndale Florida'' in Figure \ref{fig:example}  will be reformatted into ``(how) (big) (how big) (is) (big is) (how big is) (Auburndale) (is Auburndale) (big is Auburndale) (how big is Auburndale) (Florida) (Auburndale Florida) (is Auburndale Florida) (big is Auburndale Florida) (how big is Auburndale Florida)''. We can see that phrases are exhaustively detected and represented.

\begin{figure*}[t]
\setlength{\belowcaptionskip}{-10pt}
\setlength{\abovecaptionskip}{0pt}
\centering
\includegraphics[width=13cm]{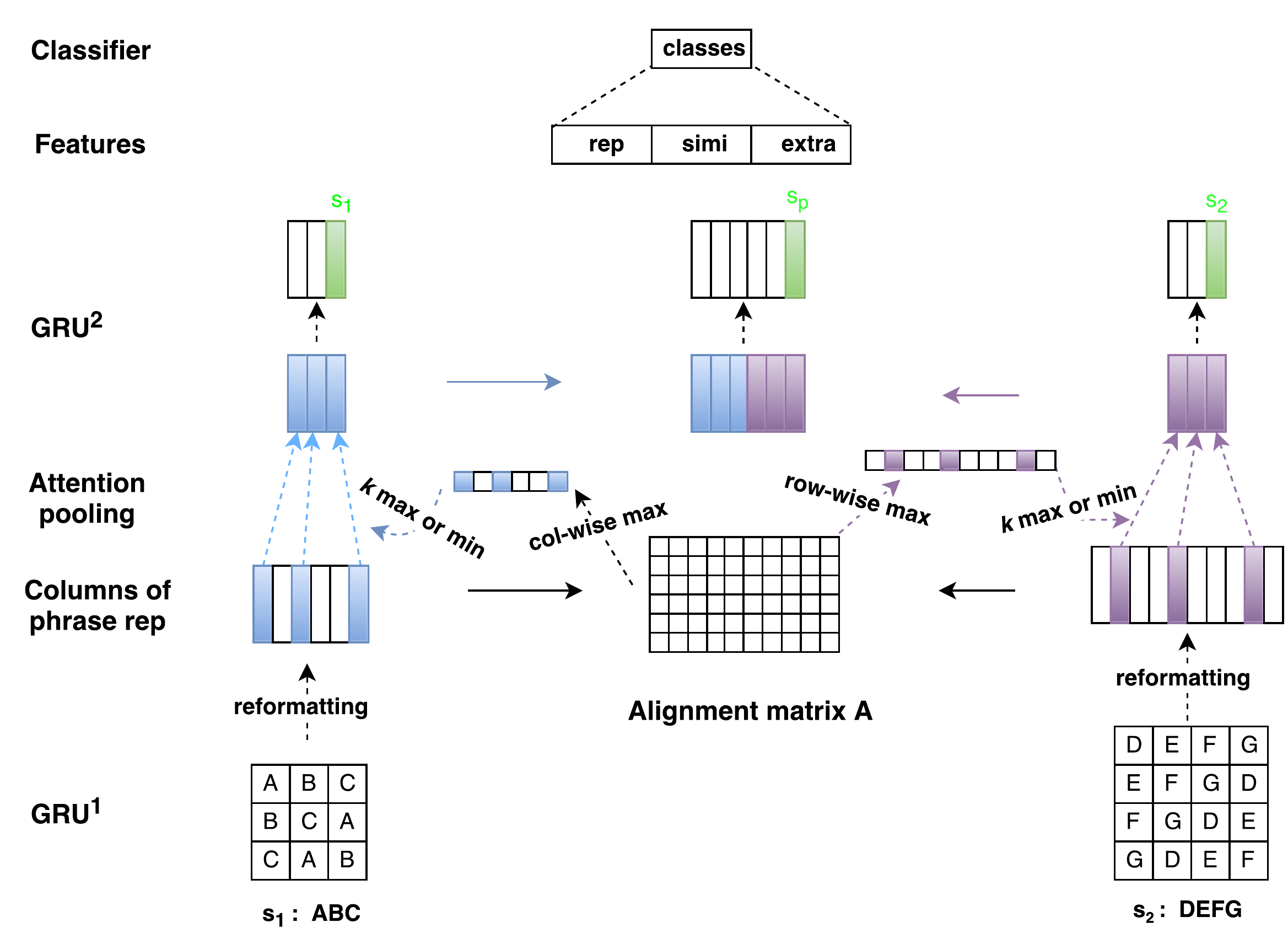}
\caption{The whole architecture} \figlabel{whole}
\end{figure*}

\subsection{Attention Pooling}\label{sec:pooling}
As each sentence $s^*$ consists of a sequence of phrases,
and each phrase is denoted by a representation vector
generated by GRU, we can compute an \emph{alignment matrix}
$\mathbf{A}$ between two sentences $s^*_1$ and $s^*_2$, by
comparing each two phrases, one from $s^*_1$ and one from
$s^*_2$. Let $s^*_1$ and $s^*_2$ also denote lengths
respectively, thus $\mathbf{A}\in\mathbb{R}^{s^*_1\times
  s^*_2}$. While there are many ways of computing the
entries of $\mathbf{A}$,  we found
that cosine works well in our setting.

The first step then is to detect the best alignment for each phrase by leveraging $\mathbf{A}$. To be concrete, for sentence $s^*_1$, we do \emph{row-wise max-pooling} over $\mathbf{A}$ as attention vector $\mathbf{a_1}$:
\begin{equation}\label{equ:att}
\mathbf{a}_{1,i}=\rm{max}(\mathbf{A}[i,:])
\end{equation}
In $\mathbf{a}_1$,  the entry $\mathbf{a}_{1,i}$ denotes the best alignment for $i^{th}$ phrase in sentence $s^*_1$. Similarly, we can do \emph{column-wise max-pooling} to generate attention vector $\mathbf{a}_{2}$ for sentence $s^*_2$. 

Now, the problem is that  we need to pay most attention to the phrases aligned very well or phrases aligned badly. According to the analysis of the two examples in Figure \ref{fig:example},  the TE example
demonstrates that we need to pay more attention to weaker
alignments whereas the AS example shows that we need to pay more
attention to stronger alignments. As a result, we adopt different second step over attention vector $\mathbf{a}_i$ ($i=1,2$) for TE and AS.

For TE, in which weaker alignments are supposed to
contribute more, we do \emph{$k$-min-pooling} over
$\mathbf{a}_i$, i.e., we only keep the $k$ phrases which
are aligned worst. For the ($Q$, $C^+$) pair in TE example
of Figure \ref{fig:example}, we expect this step is able to
put most of our attention to the phrases ``the kids'', ``the
young boys'', ``near a man with a smile'' and ``and the man
is smiling nearby'' as they have relatively weaker
alignments while their relations are the indicator of the
final decision.

For AS, in which stronger alignments are determinant, we do \emph{$k$-max-pooling} over
$\mathbf{a}_i$, i.e., we only keep the
$k$ phrases which are aligned best. For the ($Q$, $C^+$)
pair in AS example of Figure \ref{fig:example}, we expect
this $k$-max-pooling is able to put most of our attention to the
phrases ``how big'' ``Auburndale Florida'', ``the city'' and
``had a population of 12,381'' as they have relatively
stronger alignments and their relations are the indicator of
the final decision.  
We keep the
original order of extracted phrases
after $k$-min/max-pooling.

\enote{hs}{i don't think it is a good investment to spend
  any effort on this a few hours before the submission
  deadline

however, i would like to note that the last paragraph is
again a great example for why your intuition about the
difference between TE and AS may be difficult to understand
for the reader

you call the alignment ``how big'' $\leftrightarrow$ ``had a
population of 12,381'' a relatively strong alignment, but it
is not clear to me at all this is true. i guess the question
is: which 3 phrases out of the 15 phrases of ``how big is
auburndale florida'' will be picked as being strongest in
relative terms? it is not clear to me at all why ``how big''
would be one of them

}

In summary, for TE, we first do  max-pooling over
alignment matrix row-wise and column-wise respectively, then do $k$-min-pooling over generated
alignment vector; we use \emph{$k$-min-max-pooling} to
denote the whole process. In contrast, we use
\emph{$k$-max-max-pooling} for AS. 
We refer to this method of using two successive min or max
pooling steps as
\emph{attention pooling}.

Our inspiration comes from the analysis of some prior
work. For TE, \newcite{yin2015abcnn} show that
considering the pairs in which overlapping tokens are
removed can give a boost. This simple trick matches our
motivation that weaker alignment should be given more
attention in TE. However, \newcite{yin2015abcnn} remove
overlapping tokens completely, potentially obscuring
complex alignment configurations. We can treat their trick as a hard way,
and ours as a soft way, as our phrases have more flexible
lengths and the existence of overlapping phrases decreases
the risk
of missing important alignments. In
addition, \newcite{yin2015abcnn} use the same
attention mechanism for TE and AS, which is less
optimal based on our
observations. \newcite{santos2016attentive} develop a
similar attention mechanism as \cite{yin2015abcnn} specific
for AS tasks.  In experiments, we will show the superiority
of our task-specific attention mechanism.

\subsection{The Whole Architecture}
Now, we present the whole system  in Figure \ref{fig:whole}. We take sentences $s_1$ ``ABC'' and $s_2$ ``DEFG'' as illustration. Each token, i.e., A to F, in the figure is denoted by an embedding vector, hence each sentence is represented as an order-3 tensor as input. Based on tensor-style sentence input, we have described the phrase representation learning by GRU$^1$ in Section \ref{sec:rep} and attention pooling in Section \ref{sec:pooling}.

Attention pooling generates a new feature map for each
sentence, as shown in Figure \ref{fig:whole} (the 3-column matrix as \emph{input of the} GRU$^2$), and each
column representation in the feature map denotes a key
phrase in this sentence that, based on our modeling
assumptions, should be a good basis for the correct
final decision. For instance, we expect such a
feature map to contain representations of ``the young
boys'', ``outdoors'' and ``and the man is smiling nearby''
for the sentence $Q$ in TE example of Figure
\ref{fig:example}.

Now, we do another GRU step (i.e., GRU$^2$) for: 1) the new feature map of
each sentence, to encode all the key phrases as the sentence
representation; 2) a \emph{concatenated feature map} of the two
sentence feature maps, i.e., the 6-column matrix in the middle of Figure \ref{fig:whole}, to encode all the key phrases in the
two sentences sequentially as the representation of the
sentence pair. As GRU generates a hidden state at each
position, we always choose the last hidden state as the
representation of the sentence or sentence pair. In Figure
\ref{fig:whole}, these final GRU$^2$-generated representations for sentence
$s_1$, $s_2$ and the sentence pair are depicted as green
columns: $s_1$, $s_2$ and $s_p$ respectively.

As for the input of the final \emph{logistic regression} classifier, it can be
flexible, such as representation vectors (\emph{rep}),
similarity scores between $s_1$ and $s_2$ (\emph{simi}), and
extra linguistic features (\emph{extra}). This can vary
based on the specific tasks. We give details in \secref{experiments}.

\section{Experiments}\label{sec:experiments} We test the
proposed architectures on TE and AS benchmark datasets.

\subsection{Common Setup}\label{sec:cts}
For both TE and AS, words are initialized by 300-dimensional GloVe
embeddings\footnote{\url{nlp.stanford.edu/projects/glove/}} \cite{pennington2014glove} and not changed during training.  A single randomly initialized
embedding 
is created for all
unknown words 
by uniform sampling from [-.01, .01].
We use ADAM \cite{kingma2015adam}, with a first momentum coefficient of 0.9 and a second momentum coefficient of 0.999,\footnote{Standard configuration recommended by Kingma and Ba}  $L_2$
regularization and  Diversity Regularization \cite{xie2015generalization}. \tabref{hyper} shows the values of the
hyperparameters, tuned on dev.


\def\hyperspace{0.075cm}
\begin{table}[t]
\setlength{\belowcaptionskip}{-10pt}
\setlength{\abovecaptionskip}{5pt}
\begin{center}
\setlength{\tabcolsep}{2mm}
\begin{tabular}{c|cccccc}
 &$d$& lr  & bs  & $L_2$ & $div$ & $k$\\\hline
 TE&[256,256]&.0001 & 1  & .0006 & .06 & 6\\
 AS&[50,50]&.0001 &1  & .0006 & .06 &6\\
\end{tabular}
\end{center}
\caption{Hyperparameters. $d$: dimensionality of hidden states in GRU layers; lr: learning rate;  bs: mini-batch size; $L_2$: $L_2$ normalization; $div$: diversity regularizer; $k$: $k$-min/max-pooling.}\label{tab:hyper} 
\end{table}

\textbf{Common Baselines.}  (i) \textbf{Addition}. We sum up
word embeddings element-wise to form sentence
representation, then concatenate two sentence representation
vectors ($s^0_1$, $s^0_2$) as classifier input. (ii)
\textbf{A-LSTM}. The pioneering attention based LSTM system
for a specific sentence pair classification task ``natural
language inference'' \cite{entail2016}. A-LSTM
has the same dimensionality  as our GRU system in terms of initialized
word representations and the hidden states (refer to \tabref{hyper}). (iii)
\textbf{ABCNN} \cite{yin2015abcnn}. The state-of-the-art
system in both TE and AS.
 

Based on the motivation in \secref{intro}, the main
hypothesis 
to be tested in experiments is:
$k$-min-max-pooling is superior for TE and 
$k$-max-max-pooling is superior for AS. In addition, we are interested in that if the second pooling step in attention pooling, i.e., the $k$-min/max-pooling, is more effective than a ``full-pooling'' in which we forward \emph{all} the generated phrases into the next layer.

\subsection{Textual Entailment}
SemEval 2014 Task 1 \cite{marelli2014semeval} evaluates
system predictions of textual entailment (TE) relations on
sentence pairs from the SICK dataset
\cite{marelli2014sick}. The three classes are
entailment, contradiction and neutral.  The
sizes of SICK train, dev and test sets are 4439, 495 and
4906 pairs, respectively. We choose SICK benchmark dataset so that our result is directly comparable with that of \cite{yin2015abcnn}, in which non-overlapping text are utilized explicitly to boost the performance. That trick inspires this work.


Following \newcite{lai2014illinois}, we train our final
system (after fixing of hyperparameters) on train and dev
(4,934 pairs).  Our evaluation measure is accuracy.

\begin{table}[t]
\setlength{\belowcaptionskip}{-15pt}
\setlength{\abovecaptionskip}{5pt}
\begin{center}
\setlength{\tabcolsep}{1mm}
\begin{tabular}{ll|l}
\multicolumn{2}{c|}{method} &  acc  \\ \hline\hline
\multirow{3}{*}{\rotatebox{90}{\footnotesize \begin{tabular}{c}SemEval\\ Top3\end{tabular}}}&\cite{jimenez2014unal}  & 83.1 \\
 & \cite{zhao2014ecnu} & 83.6 \\
&\cite{lai2014illinois}  & 84.6 \\\hdashline
TrRNTN & \cite{bowman2015recursive}& 76.9\\\hdashline
\multirow{2}{*}{Addition}&no extra features& 73.1\\
& plus extra features &  79.4 \\\hdashline
\multirow{2}{*}{A-LSTM}&no extra features& 78.0\\
& plus extra features &  81.7 \\\hline
ABCNN & \cite{yin2015abcnn}& 86.2\\\hline\hline
\multirow{3}{*}{GRU}&$k$-max-max-pooling& 84.9 \\
&full-pooling & 85.2\\
& $k$-min-max-pooling &  \textbf{87.1}$^*$ \\\hline
\end{tabular}
\end{center}
\caption{Results on SICK. Significant
improvement over both $k$-max-max-pooling and full-pooling is marked with $*$
(test of equal proportions, p $<$ .05). }\label{tab:sickresult} 
\end{table}

\subsubsection{Feature Vector}
The final feature vector as input of classifier contains three parts: \emph{rep, simi, extra}.

\textbf{Rep}. Totally five vectors, three are the top
sentence representation $s_1$, $s_2$ and the top sentence
pair representation $s_p$ (shown in green in \figref{whole}), two are $s^0_1$, $s^0_2$ from \emph{Addition} baseline.

\textbf{Simi}. Four similarity scores, cosine similarity and
euclidean distance between $s_1$ and $s_2$, cosine
similarity and euclidean distance between $s^0_1$ and
$s^0_2$. Euclidean distance $\parallel \cdot \parallel$ is
transformed into $1/(1+\parallel \cdot \parallel)$.

\textbf{Extra}. We include the same 22 linguistic features
as \newcite{yin2015abcnn}. They cover 15 machine translation
metrics between the two sentences; whether or not the two sentences
contain negation tokens like ``no'', ``not'' etc; whether or not
they contain synonyms, hypernyms or antonyms; two sentence
lengths. See \newcite{yin2015abcnn} for details.

\begin{figure*}[!ht] 
\setlength{\belowcaptionskip}{-15pt}
\setlength{\abovecaptionskip}{0pt}
\centering 
\subfigure[Attention distribution for phrases in ``$Q$'' of TE example in Figure \ref{fig:example}] { \label{fig:vis1} 
\includegraphics[width=16cm]{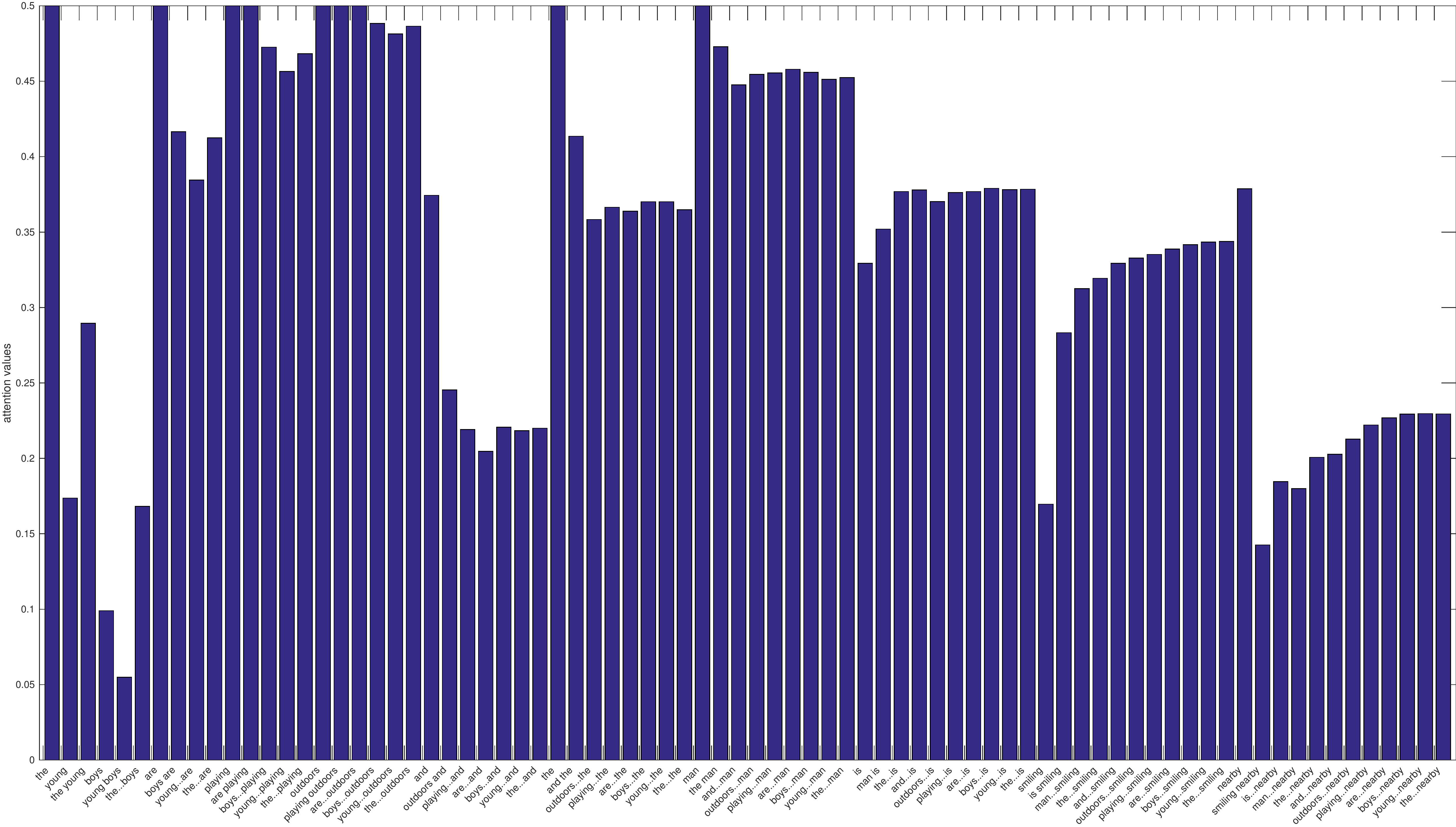} 
} 
\subfigure[Attention distribution for phrases in ``$C^+$'' of TE example in Figure \ref{fig:example}] { \label{fig:vis2} 
\includegraphics[width=16cm]{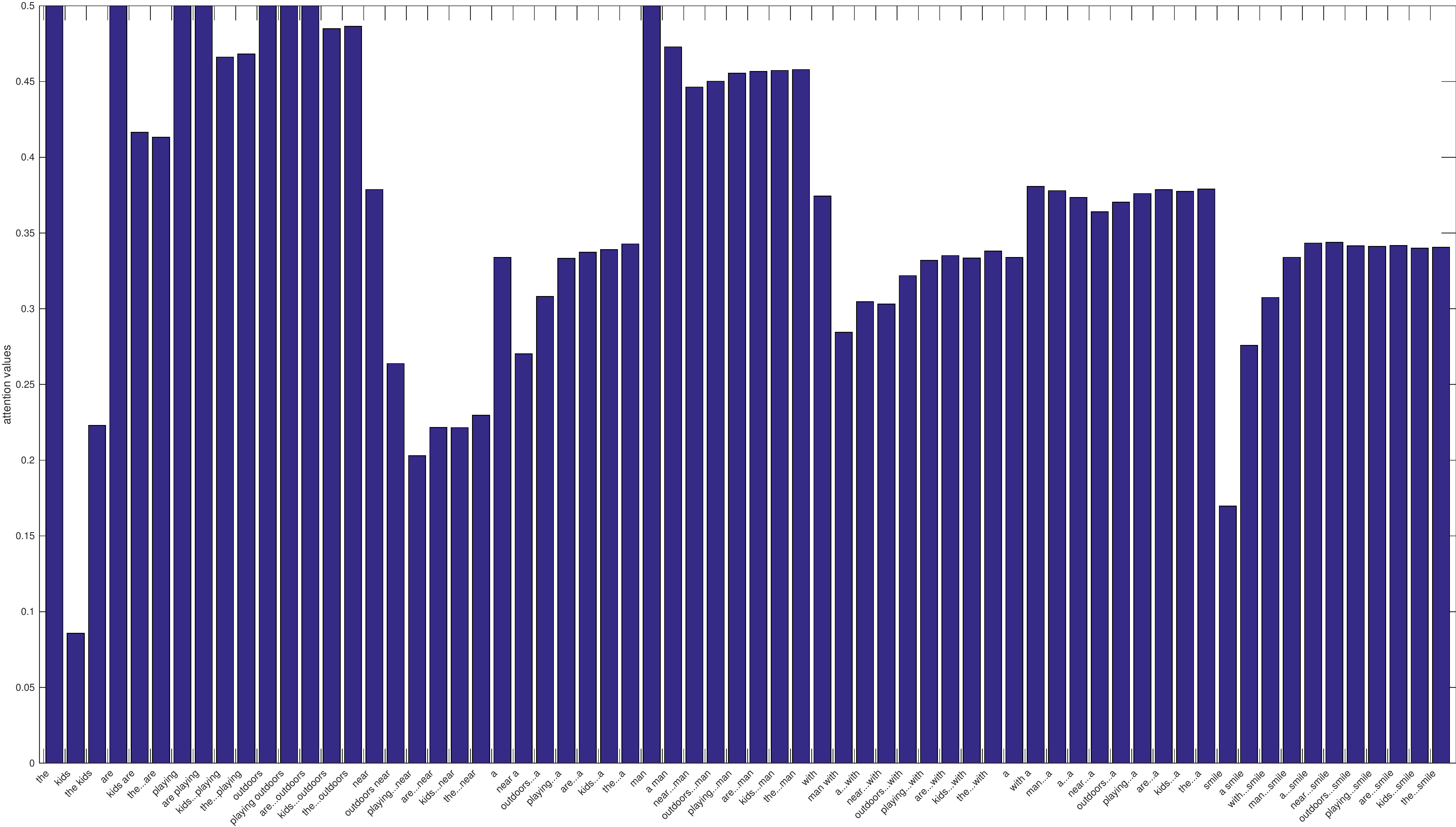} 
} 
\caption{Attention Visualization} 
\label{fig:vis} 
\end{figure*}
\subsubsection{Results}
\tabref{sickresult} shows that GRU with $k$-min-max-pooling
gets state-of-the-art performance on SICK and significantly
outperforms $k$-max-max-pooling and
full-pooling. Full-pooling has more phrase input than the
combination of $k$-max-max-pooling and $k$-min-max-pooling,
this might bring two problems: (i) noisy alignments
increase; (ii) sentence pair representation $s_p$ is no
longer discriminative, as different sentences have different
lengths, $s_p$ does not know its semantics comes from phrases
of $s_1$ or $s_2$. However, this is crucial to determine
whether $s_1$ entails $s_2$.

\enote{hs}{

didn't understand the details of this:
``sentence pair representation $s_p$ is no
longer discriminative, as different sentences have different
lengths,''

(it's clear that different lengths are a problem for full
pooling, but what does ``does not know its semantic comes
from phrases'' mean?)

you never say how exactly the input to the $s_p$ GRU is
constructed

is the input to the $s_p$ GRU in the attention pooling setup
of dimensionality $6 \times d$?

}

ABCNN \cite{yin2015abcnn} is based on assumptions similar to
$k$-max-max-pooling: words/phrases with higher matching
values should contribute more in this task. However, ABCNN
gets the optimal performance by combining a reformatted SICK
version in which overlapping tokens in two sentences are
removed. This instead hints that non-overlapping units can
do a big favor for this task, which is indeed the
superiority of our ``$k$-min-max-pooling''.

\subsection{Answer Selection}
\seclabel{as}
We use WikiQA\footnote{\url{http://aka.ms/WikiQA}
  \cite{yang2015wikiqa}}  subtask that assumes 
there is at least one correct answer for a question. 
This dataset has 20,360, 1130 and 2352 
question-candidate pairs in train, dev and test, respectively.
Results are measured by
average precision (MAP) and mean reciprocal rank (MRR).

Apart from the common baselines Addition, A-LSTM and ABCNN, we compare further with: (i) \textbf{CNN-Cnt} \cite{yang2015wikiqa}: combine CNN with
two linguistic features ``WordCnt'' (the number
of non-stopwords in the question that also occur in the
answer) and ``WgtWordCnt'' (reweight the counts
by the IDF values of the question words); (ii) \textbf{AP-CNN} \cite{santos2016attentive}.

\subsubsection{Feature Vector}
The final feature vector in AS has the same (\emph{rep, simi, extra}) structure as TE, except that \textbf{simi} consists of only two cosine similarity scores, and \textbf{extra} consists of four entries: two sentence lengths, WordCnt and
WgtWordCnt.

\begin{table}[bt]
\setlength{\belowcaptionskip}{-10pt}
\setlength{\abovecaptionskip}{0pt}
\begin{center}
\setlength{\tabcolsep}{1mm}
\begin{tabular}{ll|ll}
\multicolumn{2}{c|}{method} &  MAP  & MRR \\ \hline
\multirow{8}{*}{\rotatebox{90}{Baselines}}&CNN-Cnt & 0.6520 & 0.6652\\

&Addition& 0.5021 & 0.5069\\
&Addition (+ extra)& 0.5888 & 0.5929 \\

&A-LSTM& 0.5321 & 0.5469\\
&A-LSTM (+extra)& 0.6388 & 0.6529 \\
&AP-CNN & 0.6886 & 0.6957\\
& ABCNN & 0.6921 & 0.7127\\\hline

\multirow{3}{*}{GRU} &$k$-min-max-pooling&0.6674& 0.6791 \\
& full-pooling & 0.6693 & 0.6785\\
& $k$-max-max-pooling & \textbf{0.7108}$^*$ & \textbf{0.7203}$^*$\\\hline

\end{tabular}
\end{center}
\caption{Results on
  WikiQA\tablabel{wikiqa}. Significant
improvement over both $k$-min-max-pooling and full-pooling is marked with $*$ ($t$-test, p $<$ .05).}
\end{table}

\subsubsection{Results}
\tabref{wikiqa} shows that GRU with $k$-max-max-pooling gets the state-of-the-art results and is significantly better than its $k$-min-max-pooling and full-pooling versions. GRU with $k$-max-max-pooling has similar assumption with ABCNN \cite{yin2015abcnn} and AP-CNN \cite{santos2016attentive}: units with higher matching scores are supposed to contribute more in this task. Our improvement can be due to that: i) our linguistic units cover more exhaustive phrases, it enables alignments in a wider range; ii) we have two max-pooling steps in our attention pooling, especially the second one is able to remove some noisily aligned phrases. Both ABCNN and AP-CNN are based on convolutional layers, the phrase detection is constrained by filter sizes. Even though ABCNN tries a second CNN layer to detect bigger-granular phrases, their phrases in different CNN layers cannot be aligned directly as they are in different spaces. GRU in this work uses the same weights to learn representations of arbitrary-granular phrases, hence, all phrases can share the representations in the same space and can be compared directly.

\subsection{Visual Analysis}
In this subsection, we visualize the attention distributions over phrases, i.e., $\mathbf{a_i}$ in Equation \ref{equ:att}, of example sentences in Figure \ref{fig:example} (for space limit, we only show this for TE example). Figures \ref{fig:vis1}-\ref{fig:vis2} respectively show the attention values of each phrase in ($Q$, $C^+$) pair in TE example in Figure \ref{fig:example}. We can find that $k$-min-pooling over this distributions can indeed detect some key phrases that are supposed to determine the pair relations. Taking Figure \ref{fig:vis1} as an example, phrases ``young boys'', phrases ending with ``and'', phrases ``smiling'', ``is smiling'', ``nearby'' and a couple of phrases ending with ``nearby'' have lowest attention values. According to our $k$-min-pooling step, these phrases will be detected as key phrases. Considering further the Figure \ref{fig:vis2}, phrases ``kids'', phrases ending with ``near'', and a couple of phrases ending with ``smile'' are detected as key phrases. 

If we look at the key phrases in both sentences, we can find that the discovering of those key phrases matches our analysis in \secref{intro} for TE example: ``kids'' corresponds to ``young boys'', ``smiling nearby'' corresponds to ``near...smile''. 

Another interesting phenomenon  is that, taking Figure \ref{fig:vis2} as example, even though ``are playing outdoors'' can be well aligned as it appears in both sentences, nevertheless the visualization figures show that the attention values of ``are playing outdoors and'' in $Q$ and ``are playing outdoors near'' drop dramatically. This hints that our model can get rid of some surface matching, as the key token ``and'' or ``near'' makes the semantics of ``are playing outdoors and'' and ``are playing outdoors near'' be pretty different with their sub-phrase ``are playing outdoors''. This is important as ``and'' or ``near'' is crucial unit to connect the following key phrases ``smiling nearby'' in $Q$ or ``a smile'' in $C^+$. If we connect those key phrases sequentially as a new fake sentence, as we did in attention pooling layer of Figure \ref{fig:whole}, we can see that the fake sentence roughly ``reconstructs'' the meaning of the original sentence while it is composed of phrase-level semantic units now. 

\section{Conclusion}
This work  investigated the roles of phrase alignments of different intensities for different tasks. We argue that it is not  true that stronger alignments always contribute more. We found  TE task prefers weaker alignments while AS task prefers stronger ones.  We proposed a flexible attention pooling in GRU system to satisfy the different requirements of different tasks. Experimental results show the soundness of our argument and the effectiveness of our attention pooling based GRU systems. 



\bibliography{acl2016}
\bibliographystyle{acl2016}
\end{document}